\documentclass{article}
\usepackage{spconf,amsmath,graphicx}
\usepackage{amsfonts}
\usepackage{booktabs}

\usepackage{array,multirow,multicol,adjustbox,url}
\usepackage[dvipsnames]{xcolor}
\usepackage{hyperref}

\title{SLIM-Diff: Shared Latent Image-Mask Diffusion with Lp loss for Data-Scarce Epilepsy FLAIR MRI}

\name{
    \begin{tabular}{@{}c@{}}
    Mario Pascual-Gonz\'alez$^{\dagger}$, 
    Ariadna Jim\'enez-Partinen$^{\star \dagger \ddagger}$, 
    R.M. Luque-Baena$^{\star \dagger \ddagger}$\sthanks{Corresponding author.}, 
    F\'atima Nagib-Raya$^{\S}$, \\
    Ezequiel L\'opez-Rubio$^{\star \dagger \ddagger}$
    \end{tabular}
}

\address{
    $^{\star}$ ITIS Software, University of M\'alaga, Spain \\%, M\'alaga, 29071, Spain \\
    $^{\dagger}$ Department of Computer Languages and Computer Science, University of M\'alaga, Spain \\ %Bulevar Louis Pasteur, 35, Málaga, 29071, Spain
    $^{\ddagger}$ IBIMA Plataforma BIONAND, Instituto de Investigaci\'on Biom\'edica de M\'alaga, Spain \\
    $^{\S}$ Department of Radiology, Hospital Regional Universitario de M\'alaga, Spain
}

\begin{document}
%\ninept
%
\maketitle
\begin{abstract}
Focal cortical dysplasia (FCD) lesions in epilepsy FLAIR MRI are subtle and scarce, making joint image--mask generative modeling prone to instability and memorization. We propose SLIM-Diff, a compact joint diffusion model whose main contributions are (i) a single shared-bottleneck U-Net that enforces tight coupling between anatomy and lesion geometry from a 2-channel image+mask representation, and (ii) loss-geometry tuning via a tunable $L_p$ objective. As an internal baseline, we include the canonical DDPM-style objective ($\epsilon$-prediction with $L_2$ loss) and isolate the effect of prediction parameterization and $L_p$ geometry under a matched setup. Experiments show that $x_0$-prediction is consistently the strongest choice for joint synthesis, and that fractional sub-quadratic penalties ($L_{1.5}$) improve image fidelity while $L_2$ better preserves lesion mask morphology. Our code and model weights are available in https://github.com/MarioPasc/slim-diff 
\end{abstract}
\section{Introduction} \label{sec:introduction}

Diffusion models have enabled high-fidelity image synthesis~\cite{muller2023multimodal}, but they are typically developed and evaluated in data-rich settings. In medical imaging, and especially in rare-pathology scenarios, the number of annotated cases is inherently limited, which can destabilize diffusion training and increase the risk of memorization. This mismatch is particularly acute for focal cortical dysplasia (FCD) in epilepsy.

FCD is a leading cause of drug-resistant epilepsy, yet it remains difficult to detect automatically. These circumscribed malformations of cortical development \cite{bast2006focal} manifest on MRI as localized cortical thickening, blurring at the gray–white matter junction, and abnormal gyral or sulcal patterns \cite{najm2022ilae}, which can challenge even expert neuroradiologists \cite{widdess2006neuroimaging}. FLAIR is the sequence of choice due to its sensitivity to these abnormalities; however, FCD type I presents with a normal MRI, restricting usable cohorts to type II cases — small, single-institution case series well below the scale needed to stabilize high-capacity generative models \cite{snell2026mri}.

A promising direction is \emph{joint} image--mask synthesis, which can generate anatomically plausible images together with spatially aligned lesion annotations for data augmentation. Existing joint-synthesis frameworks primarily increase architectural capacity to model image and mask distributions. For example, MedSegFactory~\cite{mao2025medsegfactory} uses dual U-Nets with cross-attention, and brainSPADE~\cite{fernandez2022brainspade} employs a multi-stage pipeline that separates layout generation from image synthesis. These methods commonly follow the canonical DDPM training recipe with standard $\epsilon$-prediction and an $L_2$ objective~\cite{ho2020denoising}. In small-cohort epilepsy settings, however, higher-capacity designs can be harder to optimize and more prone to overfitting; moreover, the geometry of the training loss becomes consequential when lesions occupy a small fraction of pixels. Consequently, $\epsilon$-prediction with $L_2$ loss is employed as a natural internal baseline (DDPM-style objective) within our joint formulation, and then isolate the effect of prediction parameterization and loss geometry by varying only these training-objective components.

We therefore propose Shared Latent Image--Mask Diffusion (SLIM-Diff), a compact joint diffusion model designed for data-scarce epilepsy FLAIR MRI. SLIM-Diff uses a single shared-bottleneck U-Net that processes the image and mask as a coupled 2-channel input, enforcing shared representations that promote alignment while constraining capacity. Complementing this architectural inductive bias, we treat loss-geometry as an explicit design axis and evaluate tunable $L_p$ objectives with $p \in \{1.5, 2.0, 2.5\}$.

\noindent\textbf{Contributions.} The main contributions of this paper are:
\begin{itemize}
    \item A joint image--mask diffusion model based on a single shared-bottleneck U-Net, designed for data-scarce epilepsy MRI;
    \item A systematic study of tunable $L_p$ norm losses ($p \in \{1.5, 2.0, 2.5\}$) for diffusion training, showing that fractional norms provide an explicit trade-off between outlier robustness ($p < 2$) and boundary regularization ($p \geq 2$), with different optima for image and mask synthesis;
    \item An ablation study of diffusion parameterizations ($\epsilon$, $v$, $x_0$) under a matched training and evaluation protocol;
    \item A quantitative evaluation combining distributional image similarity metrics (KID/LPIPS) with mask-morphology distribution matching (MMD-MF and per-feature Wasserstein distances).
\end{itemize}

\section{Methodology}
\label{sec:methodology}

\begin{figure*}[ht]
    \centering
    \includegraphics[width=\linewidth]{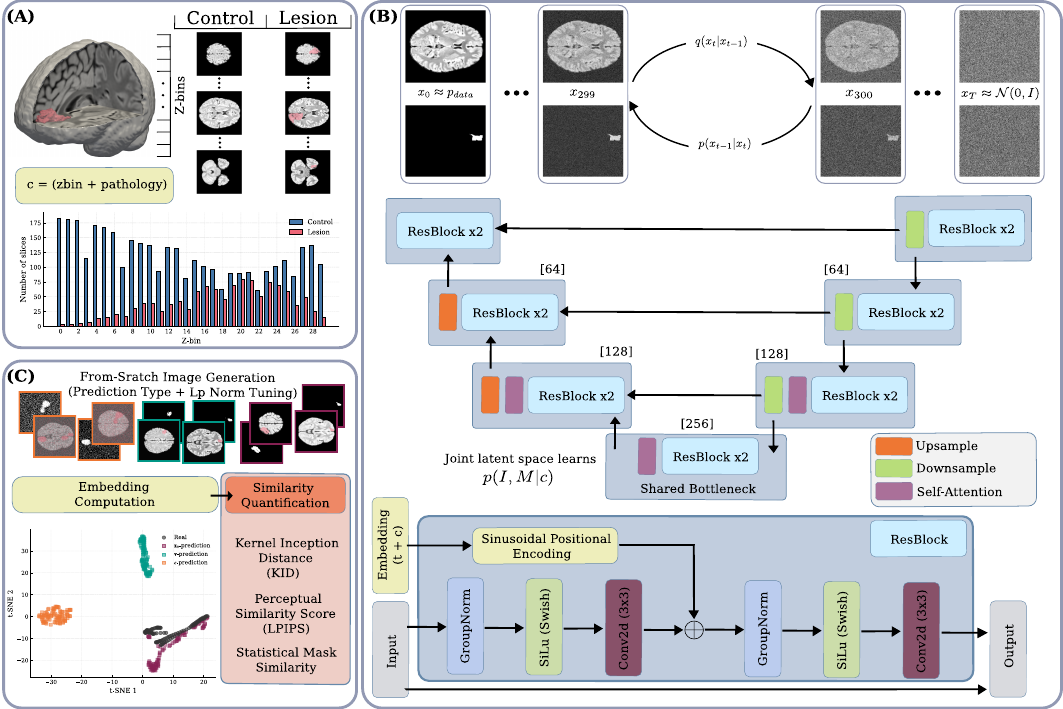}
    \caption{Overview of SLIM-Diff. (A) Training strategy. (B) Joint image--mask synthesis architecture (details in Section~\ref{sec:shared-bottleneck}) (C) Evaluation under axial-depth and pathology conditioning.}
    \label{fig:graphical-abstract}
\end{figure*}

% jsddpm-similarity-metrics plot \
%--config src/diffusion/scripts/similarity_metrics/config/icip2026_self_cond_0.0.yaml \
%--add-mask-metrics \
%--output-subdir plots_with_masks    

\begin{figure*}[hbtp]
    \centering
    \includegraphics[width=0.93\linewidth]{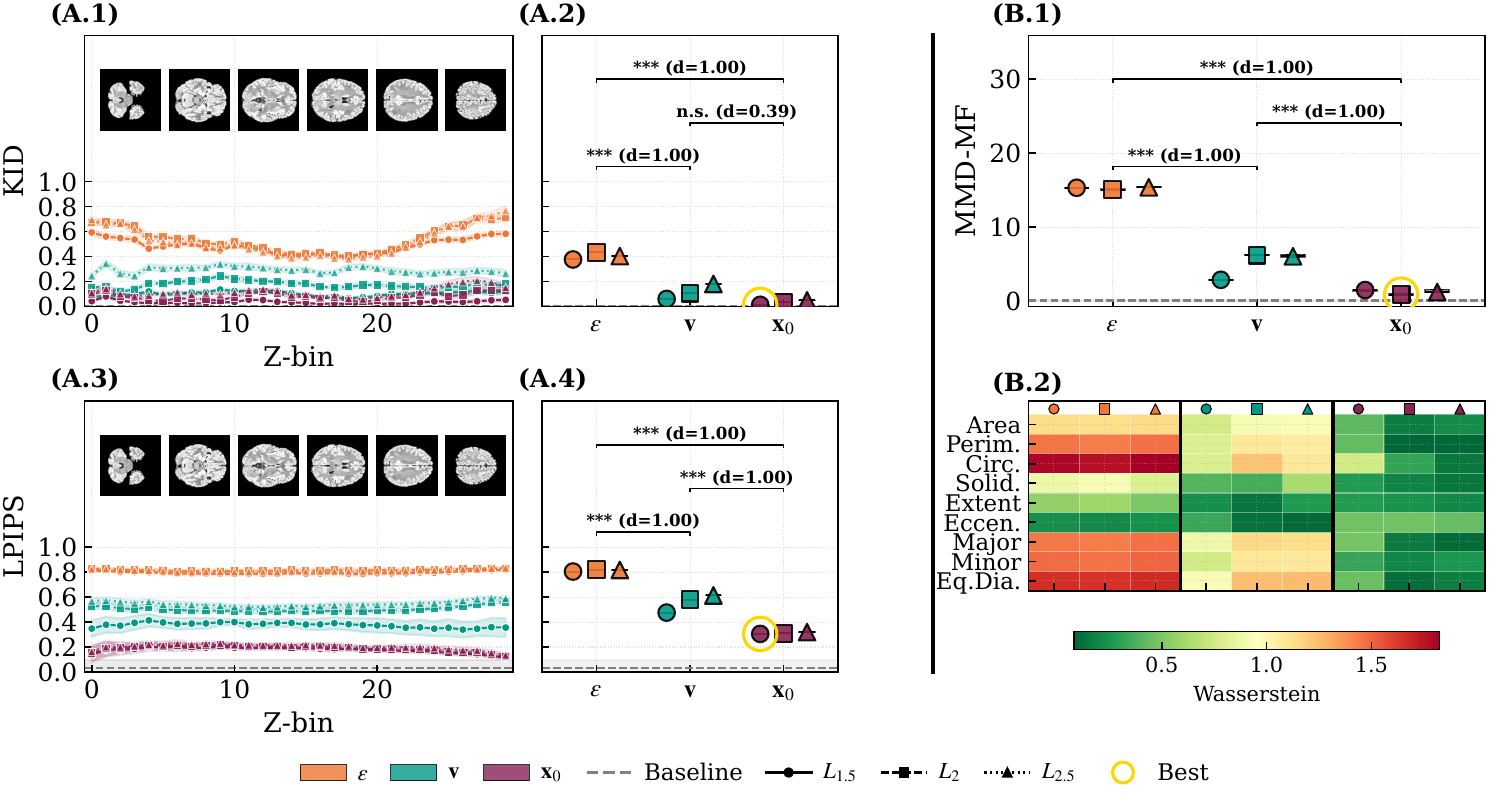}
    \caption{Similarity and mask-quality metrics for SLIM-Diff across prediction targets ($\epsilon$, $v$, $x_0$) and $L_p$ settings. (A) Image realism is quantified with KID and LPIPS (lower is better), reported against a held-out real test set and contextualized with a real-vs-real baseline (two disjoint subsets of real data under the same protocol). (B) Lesion mask realism is quantified with MMD-MF (Maximum Mean Discrepancy on Morphological Features) and complemented with per-feature Wasserstein distances over nine standard shape descriptors; the best configuration is highlighted in the figure.}
    \label{fig:p1}
\end{figure*}
% \vspace{-0.75\baselineskip}

 % \begin{figure}[hbtp]
 %     \centering
 %     \includegraphics[width=\linewidth]{image_grid_publication.pdf}
 %     \caption{Qualitative joint synthesis results from SLIM-Diff. Each example shows a generated FLAIR slice and its jointly generated lesion mask under the same conditioning token (axial $z$-bin and pathology class), illustrating spatial alignment between anatomy and lesion location across slice depths.}
 %     \label{fig:p2}
 % \end{figure}

\subsection{Dataset}
\label{sec:dataset}

Schuch et al. \cite{schuch2023open} created an open-access annotated dataset with 85 epilepsy patients -- 78 suspected FCDII, 5 MRI-negative, and 2 with other abnormalities -- and 85 healthy controls. For the proposed method, to avoid a large imbalance and potential bias, only the 78 FLAIR sequences corresponding to FCDII were used. The MRI scans provided are intra-subject registered and defaced.  Raw MRI data undergo preprocessing to ensure intra- and inter-subject homogeneity. This involves: registration to the $1 mm^3$ MNI152 space with the ``SyN'' (affine and deformable transformation), skull-stripping to remove non-brain tissues \cite{tustison2021antsx}, bias correction of inhomogeneous intensities from the same tissue applying the N4 bias field correction algorithm \cite{tustison2010n4itk}. Finally, 3D MRI scans were reduced to a resolution of 1.25mm and split into 2D slices from the axial plane. The control (non-lesion) images were obtained from areas without lesions. The number of control and lesion slices along the z-axis is illustrated in panel A of \figurename~\ref{fig:graphical-abstract}.

\subsection{Problem Formulation}
\label{sec:problem-formulation}

Limited cohort size and lesion heterogeneity demand a constrained yet effective model architecture. Given a conditioning signal $c = (z_{\text{bin}}, p)$ encoding axial position and pathology class, we model the joint distribution:
\begin{equation}
p_\theta(I, M \mid c) \quad \text{where} \quad I \in \mathbb{R}^{H \times W}, \; M \in \{-1, +1\}^{H \times W}
\label{eq:joint-distribution}
\end{equation}
Our procedure directly parametrizes the joint distribution through a single neural network predicting both modalities simultaneously.

\subsection{Joint-Synthesis Diffusion}
\label{sec:joint-synthesis-diffusion}

Figure~\ref{fig:graphical-abstract} depicts the proposed approach. Joint-Synthesis Diffusion stochastically binds image and mask into a unified forward-reverse process. We define the joint sample as $\mathbf{x}_0 = [I, M]^\top \in \mathbb{R}^{2 \times H \times W}$, with both channels -- FLAIR slice and mask -- normalized to $[-1, 1]$ via percentile-based intensity scaling (0.05th and 99.5th percentiles). The forward diffusion follows:
\begin{equation}
q(\mathbf{x}_t \mid \mathbf{x}_0) = \mathcal{N}\left(\mathbf{x}_t; \sqrt{\bar{\alpha}_t} \mathbf{x}_0, (1 - \bar{\alpha}_t) \mathbf{I}\right)
\label{eq:forward-diffusion}
\end{equation}
where $\bar{\alpha}_t = \prod_{s=1}^{t} \alpha_s$ follows a cosine schedule~\cite{nichol2021improved} over $T = 1000$ timesteps. An identical stochastic corruption was applied to both channels to preserve spatial alignment throughout the diffusion trajectory: using the same noise realization for image and mask ensures corresponding spatial locations remain coupled at every timestep, which encourages the reverse model to recover anatomically consistent image-mask pairs. During the reverse process $p_\theta(\mathbf{x}_{t-1} \mid \mathbf{x}_t, c)$, regions predicted as lesion in the mask channel bias the image channel toward lesion-consistent intensities, exploiting the intrinsic image-mask coupling.

\subsection{Proposed Architecture: Shared Bottleneck UNet}
\label{sec:shared-bottleneck}

Prior joint-synthesis approaches often scale model complexity (e.g., by coupling parallel U-Nets with cross-attention or using multi-stage pipelines). We instead propose an information bottleneck that forces the model to learn $p(I, M \mid c)$ via shared convolutional features, minimizing parameter count and overfitting risk while maintaining generalization.

The architecture employs a single 2-channel UNet operating on $160 \times 160$ slices, with channel progression [64, 128, 256, 256], multi-head self-attention (32 channels per head) restricted to the two deepest levels, and 2 residual blocks per level with GroupNorm (32 groups). This constrained capacity (compared to 320–1280 channels in Stable Diffusion~\cite{rombach2022high}) serves as implicit regularization. The shared bottleneck at $256 \times 20 \times 20$ forces the network to discover latent factors explaining both FLAIR hyperintensity patterns and mask geometry, rather than memorizing sample-specific correlations.

\subsection{Conditioning Mechanism}
\label{sec:conditioning-mechanism}

We discretize the axial MRI dimension into $N_z = 30$ bins, accepting that anatomically similar slices fall into the same bin. Combined with pathology class $p \in \{0, 1\}$, this yields 60 unique conditions:
\begin{equation}
\text{token} = z_{\text{bin}} + p \cdot N_z, \quad z_{\text{bin}} \in \{0, \ldots, 29\}
\label{eq:token}
\end{equation}
The conditioning embedding combines learned pathology representations with sinusoidal z-position encoding for smooth interpolation between bins:
\begin{equation}
\mathbf{c}_{\text{emb}} = \text{Linear}\left( \left[ \mathbf{E}_p[p] \; \| \; \text{SinPE}(z_{\text{bin}}) \right] \right)
\label{eq:class-embedding}
\end{equation}
where $\mathbf{E}_p \in \mathbb{R}^{2 \times d}$ is a learned pathology embedding and $\text{SinPE}(\cdot)$ denotes sinusoidal positional encoding with geometric frequency progression $\omega_i = 1/10000^{2i/d}$. This fixed encoding provides an inductive bias for spatial continuity, enabling generalization to underrepresented z-positions.

The timestep embedding similarly employs sinusoidal encoding followed by a 2-layer MLP:
\begin{equation}
\mathbf{t}_{\text{emb}} = \text{MLP}\left( \text{SinPE}(t) \right)
\label{eq:time-embedding}
\end{equation}
The final embedding $\mathbf{e} = \mathbf{t}_{\text{emb}} + \mathbf{c}_{\text{emb}}$ is injected into each ResBlock via additive modulation after the first convolution, as shown in \figurename~\ref{fig:graphical-abstract}.

\subsection{Proposed Training Strategy: Lp loss}
\label{sec:training-strategy}

We employ fully conditioned training; an unconditional model tends to learn a dataset average that may not correspond to anatomically plausible samples. To address class imbalance, we perform subject-level oversampling of lesion cases so the model sees lesion and non-lesion subjects equally often during training; this is done after the subject-wise train/val split, so it does not introduce leakage.

Standard diffusion models minimize an $L_2$ loss derived from the variational bound~\cite{ho2020denoising}. We adopt this canonical recipe as an internal baseline and keep architecture and conditioning fixed so that observed differences arise from training-objective choices. We then use a tunable $L_p$ norm and compare prediction targets (noise, velocity~\cite{salimans2022progressive}, and $x_0$), as specified above.

Concretely, our training objective is:
\begin{equation}
\mathcal{L} = \mathbb{E}_{t, \mathbf{x}_0, \boldsymbol{\epsilon}} \left[ \| \text{target} - f_\theta(\mathbf{x}_t, t, c) \|_p^p \right]
\label{eq:loss}
\end{equation}
where the target depends on the chosen prediction type and the exponent $p$ is swept over the values introduced in Section~\ref{sec:introduction}. We hypothesize this interaction is non-trivial: $\epsilon$-prediction targets unit-variance Gaussian noise regardless of data statistics, whereas $x_0$-prediction targets the empirical data distribution.

Training employs AdamW optimization with learning rate $10^{-4}$, cosine annealing schedule, and gradient clipping at norm 1.0. Exponential moving average (EMA) of weights with decay 0.999 is maintained for inference. Early stopping with patience of 25 epochs monitors validation loss. Inference uses fully conditioned DDIM sampling~\cite{song2020denoising} with 300 steps and stochasticity $\eta = 0.2$; generation always requires a conditioning token (Section~\ref{sec:conditioning-mechanism}), i.e., we do not sample unconditionally.

\subsection{Evaluation Protocol}
\label{sec:evaluation-protocol}

\noindent\textbf{Metrics.} We evaluate joint image--mask synthesis using (i) Kernel Inception Distance (KID)~\cite{binkowski2018demystifying} for distributional image realism in Inception feature space, (ii) Learned Perceptual Image Patch Similarity (LPIPS)~\cite{zhang2018perceptual} for perceptual similarity to held-out real samples, and (iii) mask realism via Maximum Mean Discrepancy on Morphological Features (MMD-MF) computed on nine standard lesion-shape descriptors (area, perimeter, circularity, solidity, extent, eccentricity, major/minor axis length, equivalent diameter). For interpretability, we also report a real-vs-real baseline computed between two disjoint subsets of real data under the same protocol.

%\noindent\textbf{Computation and uncertainty.} KID and MMD-MF are computed as an unbiased MMD$^2$ estimator with a degree-3 polynomial kernel, reported as $mean\pm std$ over 100 random subsets (subset sizes: 1000 for image features, 500 for mask features). For MMD-MF, morphology features are Z-score normalized using statistics from real data only. LPIPS is reported as $mean\pm std$ over random real--synthetic pairings.

\noindent\textbf{Statistical testing.} We use non-parametric tests ($\alpha=0.05$). To compare prediction types ($\epsilon$, $v$, $x_0$), we run a Kruskal--Wallis H-test pooling all replicas (independent trainings) across all $L_p$ settings; when significant, we apply Dunn's post-hoc test with Benjamini--Hochberg FDR correction. Within each prediction type, we compare $L_p\in\{1.5,2.0,2.5\}$ using a Friedman test (repeated measures over replicas); when significant, we apply a Nemenyi post-hoc test. We report Cliff's delta $\delta$ alongside $p$-values.

\section{Discussion}
\label{sec:discussion}

Table~\ref{tab:similarity_metrics} and \figurename~\ref{fig:p1} summarize our main quantitative results, reporting image realism (KID/LPIPS) and lesion mask realism (MMD-MF and per-feature Wasserstein distances) across prediction targets ($\epsilon$, $v$, $x_0$) and $L_p$ settings. We use them as the primary evidence for the following discussion.

% ILLUSTRATIONS & COLOR: Illustrations must appear within the designated margins. They may span the two columns. If possible, position illustrations at the top of columns, rather than in the middle or at the bottom. Caption and number every illustration. All halftone illustrations must be clear in black and white. Since the printed proceedings will be produced in black and white, be sure that your images are acceptable when printed in black and white (the electronic, conference-distributed proceedings and the IEEE Xplore proceedings will retain the colors in your document)
\begin{table}[htbp]
  \centering
  \caption{Global similarity metrics by prediction type and $L_p$ norm (lower is better). Best values per metric are highlighted in \textbf{Bold}, while worst values are \underline{Underlined}. As mentioned in Section~\ref{sec:introduction}, our adopted internal baseline consists of the $\epsilon$-prediction and $L_{p=2.0}$. }
  \label{tab:similarity_metrics}
  \begin{tabular}{llccc}
    \toprule
    P & $L_p$ & KID $\downarrow$ & LPIPS $\downarrow$ & MMD-MF $\downarrow$ \\
    \midrule
    $\epsilon$ & 1.5 & 0.376 $\pm$ 0.001 & 0.805 $\pm$ 0.004 & 15.34 $\pm$ 0.08 \\
     & 2.0 & \underline{0.432 $\pm$ 0.002} & \underline{0.821 $\pm$ 0.002} & 15.06 $\pm$ 0.12 \\
     & 2.5 & 0.403 $\pm$ 0.001 & 0.817 $\pm$ 0.001 & \underline{15.40 $\pm$ 0.02} \\
    \midrule
    $\mathbf{v}$ & 1.5 & 0.059 $\pm$ 0.002 & 0.475 $\pm$ 0.001 & 2.86 $\pm$ 0.06 \\
     & 2.0 & 0.105 $\pm$ 0.001 & 0.579 $\pm$ 0.003 & 6.25 $\pm$ 0.13 \\
     & 2.5 & 0.180 $\pm$ 0.001 & 0.613 $\pm$ 0.007 & 6.16 $\pm$ 0.19 \\
    \midrule
    $\mathbf{x}_0$ & 1.5 & \textbf{0.012 $\pm$ 0.001} & \textbf{0.305 $\pm$ 0.000} & 1.43 $\pm$ 0.22 \\
     & 2.0 & 0.034 $\pm$ 0.000 & 0.310 $\pm$ 0.002 & \textbf{0.95 $\pm$ 0.13} \\
     & 2.5 & 0.048 $\pm$ 0.000 & 0.319 $\pm$ 0.005 & 1.31 $\pm$ 0.29 \\
    \bottomrule
  \end{tabular}
\end{table}
\vspace{-0.75\baselineskip}

Across our experiments, $x_0$-prediction performs best in this setup (KID, LPIPS, MMD-MF; $p < 0.001$, Cliff's $\delta = 1.0$). A plausible explanation is that predicting $x_0$ provides a more structured training signal than $\epsilon$-prediction, which targets Gaussian noise. This difference may matter more in data-scarce regimes: the effective variance of an $x_0$ target is tied to the empirical data distribution, whereas $\epsilon$ maintains unit variance regardless of dataset size, which can translate into higher-variance gradients. The velocity parameterization ($v$-prediction)~\cite{salimans2022progressive} yields intermediate performance, consistent with it interpolating between $x_0$ and $\epsilon$ targets.

This effect is amplified by the shared-bottleneck architecture, which constrains capacity by forcing image and mask information through a shared representation. Our U-Net has 26.9M parameters, substantially smaller than common large diffusion backbones (e.g., the 860M-parameter U-Net used in Stable Diffusion) and smaller than dual-stream joint-synthesis designs that maintain separate networks for image and mask. This compactness is a pragmatic choice for the low-data regime; empirically, training was stable, and early stopping (patience=25) helped limit overfitting.

The divergence between optimal $L_p$ norms for image quality ($L_{1.5}$) versus mask morphology ($L_{2.0}$) is a key empirical finding. We interpret this through the lens of robust statistics: sub-quadratic penalties ($p < 2$) reduce the influence of high-residual pixels~\cite{barron2019general}, which in FLAIR reconstruction often correspond to lesion boundaries and hyperintense regions that deviate maximally from background tissue. By down-weighting these ``outliers,'' $L_{1.5}$ better preserves subtle intensity gradients without over-penalizing anatomically meaningful deviations. Conversely, mask synthesis benefits from quadratic loss: binary boundaries require precise localization where any pixel error is penalized uniformly, favoring the mean-seeking behavior of $L_2$.

This task-dependent optimum suggests that $L_p$ tuning provides a complementary regularization axis orthogonal to architectural constraints: while the shared bottleneck limits capacity, $p$ shapes the loss landscape geometry. The per-feature Wasserstein breakdown indicates that matching coarse lesion size statistics is generally easier than matching fine-grained shape descriptors. %We additionally observe degraded performance at extreme z-bins (0--2, 27--29), where training samples are sparse (Fig.~\ref{fig:graphical-abstract}).

\section{Conclusion}
\label{sec:conclusion}

We presented SLIM-Diff, a shared-bottleneck diffusion framework for joint FLAIR image and lesion mask synthesis in data-scarce epilepsy imaging. Our ablation indicates that $x_0$-prediction is the most effective parameterization in this setting, and that tunable $L_p$ losses provide task-specific optimization: sub-quadratic $L_{1.5}$ improves image quality while $L_2$ better preserves lesion mask morphology. These results suggest that loss design deserves attention commensurate with architectural constraints, particularly in low-data medical regimes.

\subsection{Limitations and Future Work}
\label{sec:limitations-and-future-work}

The most significant constraint of our current framework is its reliance on 2D slice-based generation, which prioritizes sample efficiency over explicit volumetric consistency. While this design choice effectively mitigates the severe data scarcity inherent to epilepsy datasets, it risks introducing subtle anatomical discontinuities across the z-axis if the intent is to generate full-brain data. However, recent work in low-data regimes suggests that 2D approaches can match or even outperform 3D counterparts when training data is limited or anisotropic~\cite{isensee2021nnu}. To bridge this gap, future iterations should explore pseudo-3D consistency mechanisms, such as slice-to-volume self-labelling or orthogonal plane conditioning, to enforce anatomical continuity without the prohibitive computational cost of full volumetric diffusion.

% Beyond architecture, our reliance on distributional metrics (like KID and LPIPS) and geometric statistics quantifies perceptual quality but does not guarantee that synthetic samples will actually improve diagnostic performance. To address this, our immediate next step is to integrate generated samples into state-of-the-art segmentation pipelines. By measuring improvements in sensitivity and False Positive Rates on held-out test sets, we can directly assess the clinical utility of our synthetic augmentation strategies. 

Finally, the current work has been validated primarily through internal ablation, leaving its relative standing against other joint-synthesis frameworks undefined. We have not performed direct comparisons with dual-stream architectures such as MedSegFactory~\cite{mao2025medsegfactory} or multi-stage pipelines like brainSPADE~\cite{fernandez2022brainspade}. A principled baseline comparison requires matched conditioning and domain: for example, MedSegFactory's released pre-trained weights were trained on non-neuroimaging datasets and do not include the z-position conditioning central to our slice-level generation. brainSPADE, in turn, follows a two-stage paradigm (label synthesis followed by image generation) and requires multi-class one-hot tissue labels rather than binary lesion masks, and does not provide publicly available pre-trained weights, making a faithful adaptation to FCD/FLAIR non-trivial within our evaluation timeframe. Running such methods without domain-appropriate fine-tuning/adaptation would therefore yield uninformative comparisons. We leave cross-architecture benchmarking under matched low-data epilepsy conditions to future work.

\section*{Acknowledgements}
This work is partially supported by the Autonomous Government of Andalusia (Spain) under project UMA20-FEDERJA-108, and also by the Ministry of Science and Innovation of Spain, grant number PID2022-136764OA-I00. It includes funds from the European Regional Development Fund (ERDF). It is also partially supported by the Fundaci\'on Unicaja under project PUNI-003\_2023, the Instituto de Investigaci\'on Biom\'edica de M\'alaga y Plataforma en Nanomedicina-IBIMA Plataforma BIONAND under project ATECH-25-02, and the Instituto de Salud Carlos III, project code PI25/02129 (co-financed by the European Union). The authors thankfully acknowledge the computer resources, technical expertise and assistance provided by the SCBI (Supercomputing and Bioinformatics) center of the University of M\'alaga.

\bibliographystyle{IEEEbib}
\bibliography{strings,refs}

\end{document}